\documentclass[10pt, a4paper, onecolumn]{article}

\usepackage[numbers]{natbib}

\usepackage[T1]{fontenc}
\usepackage{lmodern}

\usepackage{natbib}
\usepackage[utf8]{inputenc} 
\usepackage{booktabs}       
\usepackage{amsfonts}       
\usepackage{nicefrac}       
\usepackage{microtype}      
\usepackage{xcolor}         
\usepackage{graphicx}
\usepackage{amsmath,amssymb,amsfonts}
\usepackage{bm}
\usepackage{url}

\usepackage{array,multirow,graphicx}

\usepackage{geometry}
\newgeometry{vmargin={15mm}, hmargin={12mm,17mm}} 

\title{Normalized Validity Scores for DNNs in Regression based Eye Feature Extraction}

\author{
	Wolfgang Fuhl
	University Tübingen\\
	Tübingen, 72076 \\
	\texttt{wolfgang.fuhl@uni-tuebingen.de} \\
}

\begin{document}
	
	\maketitle
	
	\begin{abstract}
		We propose an improvement to the landmark validity loss. Landmark detection is widely used in head pose estimation, eyelid shape extraction, as well as pupil and iris segmentation. There are numerous additional applications where landmark detection is used to estimate the shape of complex objects. One part of this process is the accurate and fine-grained detection of the shape. The other part is the validity or inaccuracy per landmark, which can be used to detect unreliable areas, where the shape possibly does not fit, and to improve the accuracy of the entire shape extraction by excluding inaccurate landmarks. We propose a normalization in the loss formulation, which improves the accuracy of the entire approach due to the numerical balance of the normalized inaccuracy. In addition, we propose a margin for the inaccuracy to reduce the impact of gradients, which are produced by negligible errors close to the ground truth.
	\end{abstract}

	\section{Introduction}
	\begin{figure}
		\centering
		\includegraphics[width=\textwidth]{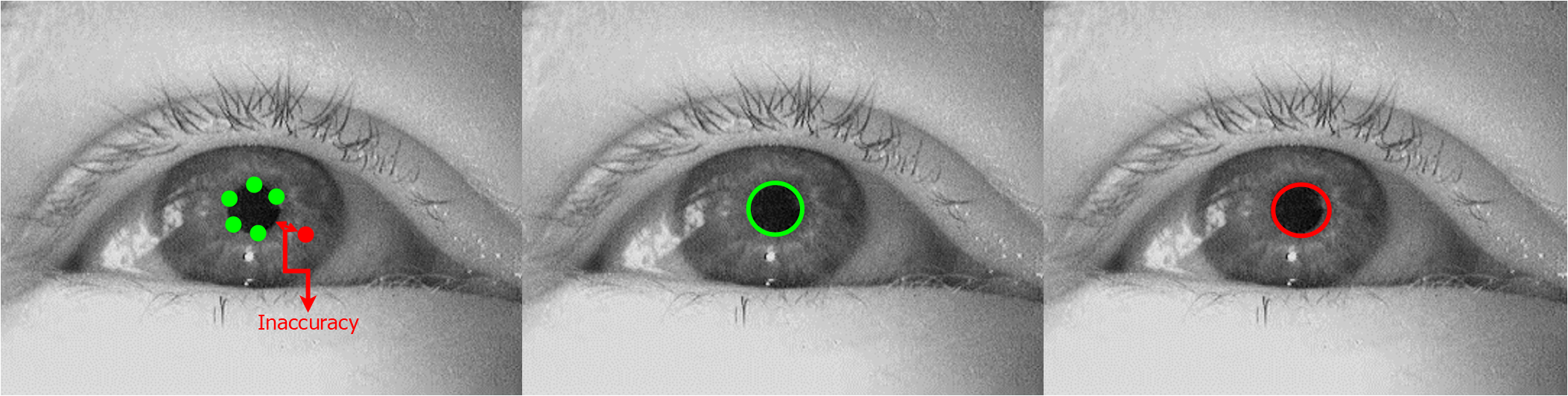}
		\caption{Impact of inaccurate landmarks on the final shape of the extracted object.}
		\label{fig:teaser}
	\end{figure}
	
	The early approaches for gaze estimation stem from landmark detection by detecting the pupil center~\cite{duchowski2002breadth,holmqvist2023eye}. A multitude of approaches have been proposed to solve this problem~\cite{morimoto2000pupil,fuhl2016pupilnet,fuhl2015excuse,fuhl2016else,santini2018pure,vera2019deepeye,fuhl2022pistol}, and based upon those detected landmarks, a multitude of application areas have risen in many different disciplines like, medicine~\cite{harezlak2018application,ashraf2018eye}, virtual reality~\cite{clay2019eye,rappa2022use}, manufactoring~\cite{zheng2022opportunities,mehta2020mining}, gaze-based behaviour analysis~\cite{gidlof2013using,muthumanickam2016supporting}, and many more. Landmark localization refers to the process of determining the precise location of specific points on an object within an image. These points are typically identified by their spatial relationship to one another, such as the nose and eyes on a face. Even when the object undergoes changes in pose, the landmarks do not randomly shift position, but rather maintain their relative positions to each other. Landmark localization is a common challenge in various applications, including 2D face alignment, face reconstruction, and gesture recognition \cite{1,2}. It also serves as a crucial step in tasks like head pose estimation (3, 4), emotion estimation \cite{5,6}, and face recognition \cite{7,8}. The current state of the art in this field focuses on addressing the challenges of accurate and robust landmark detection in both real-world and controlled environments. Deep neural networks have gained significant attention in recent years as they have outperformed other machine learning approaches in various computer vision tasks. This includes landmark detection, where different network architectures such as convolutional neural networks \cite{12}, autoencoders \cite{13}, recurrent networks \cite{14}, residual networks \cite{15,16}, and hourglass networks \cite{1} have been employed \cite{9,10,11}. In recent developments, the significance of the loss function as a key element for enhancing the accuracy and robustness of landmark localization has been emphasized. One notable example is the Wing loss, proposed to improve performance \cite{15}. Another innovative architecture, known as the local to global context network, has been introduced to output heat map candidates and effectively detect landmarks for multiple individuals in a single image \cite{17}. These advancements contribute to the continuous progress in achieving precise and reliable landmark localization in various applications. 
	
	All modern approaches mentioned so far deal with the detection problem itself, but none of these methods is capable of detecting its own inaccuracy, e.g., the validity of the detected landmarks. The approach proposed in \cite{fuhl2020learning} is the first that enables the neural network to estimate its own failure. Our work builds upon this approach by additionally estimating the inaccuracy of the landmarks. We extended the loss formulation with a normalization factor to balance the loss produced by the inaccuracy with the loss of the landmarks. In addition, we integrate a margin approach that removes tiny gradients if an inaccuracy estimate is close to the real inaccuracy.

	In short, our contributions are:
	\begin{itemize}
		\item An extended equation for the joint landmark inaccuracy loss
		\item A margin approach to exclude tiny gradients for the joint landmark inaccuracy loss
		\item Evaluations for the impact on the accuracy and the usefulness of the inaccuracy for correction.
	\end{itemize}

	\section{Related work}
	Landmark detection using deep learning relies on regression, meaning that the network directly predicts the landmark locations. We classify these methods by their architecture, data balancing, cascading, and loss functions. A simple way to apply a CNN for 2D landmark regression was suggested in ~\cite{12, 15}. The network takes a face image as input, which is obtained from a prior face detection step. New architectures like residual ~\cite{15} and recurrent networks ~\cite{14} have enhanced the performance of regression-based landmark detection. Hourglass networks ~\cite{1} have also improved the accuracy and robustness of landmark detection. However, they produce a heat map as output, where each pixel indicates the likelihood of being a landmark location, rather than a position vector. Facial landmarks are affected by large pose variations. However, most of the current datasets have a strong bias towards frontal faces; extreme angles are seldom present in the data. To address this issue, multiview models were introduced, dividing the problem into frontal and profile faces. These were already employed in conventional methods, such as ASM ~\cite{18} and AAM ~\cite{19}, as well as for cascaded regression-based methods ~\cite{20, 21}. In ~\cite{16}, a cycle GAN was used to create images to deal with style variations.
	
	In eye tracking there are also approaches which rely on landmark detection. Pistol~\cite{fuhl2022pistol} for example uses specially designed deep learning models with the recession task for landmark estimation. This is done for the pupil, the iris, and the eyelids. For the pupil and the Iris the landmarks are placed along the couture. The eyelid estimation is done similarly, but here the contour is based on a Bézier curve which is equally separated into landmarks. For the large TEyeDS dataset~\cite{fuhl2021teyed}, the authors also employed the contour separation into landmarks for the detection models. In contrast to Pistol, the authors from TEyeDS used a polynomial for the eyelid estimation. Classical approaches like ElSe~\cite{fuhl2016else}, ExCuSe~\cite{fuhl2015excuse}, PuRe~\cite{santini2018pure} etc. perform also landmark detection but follow classical algorithms from computer vision which is mainly the edge detection and a refinement of the edges for final pupil center estimation. There exist also U-Net~\cite{ronneberger2015u} approaches for segmentation like EllSeg~\cite{kothari2021ellseg} or DeepVOG~\cite{yiu2019deepvog} but they perform pixel wise classification and no landmark detection which is why they are not part of this work.
	
	In this paper, we improve the approach from \cite{fuhl2020learning} for landmark inaccuracy estimation. The original approach used the absolute error without any normalization and is shown in Equation~\ref{eq:Original}.
	\begin{equation}
		Loss_{Inaccuracy}=((\sum_{i=0}^{GroupSize} |GT_{i}-ES_{i}|) - ES_{Inaccuracy})^2
		\label{eq:Original}
	\end{equation}
	The loss function itself is the mean-squared error, and the inaccuracy is estimated over all neural outputs which belong to the same landmark (A Group). For this, the authors computed the absolute error of the landmark estimates and let an additional neural output estimate this error. The problem with this approach is that the inaccuracy estimate is numerically larger compared to the landmark estimates. This leads to an overweighting of the gradient and therefore a lower performance, as it is shown in this paper. In addition, we included a margin in the loss formulation to avoid tiny gradients influencing the learning of the deep neural network. Those tiny gradient usually come from the numerical inaccuracy of the floating point representation in modern computers, as well as from inaccurately annotated landmarks. In the following, we will describe our approach in detail. 
	
	\section{Method}
	
	\begin{figure}
		\centering
		\includegraphics[width = \textwidth]{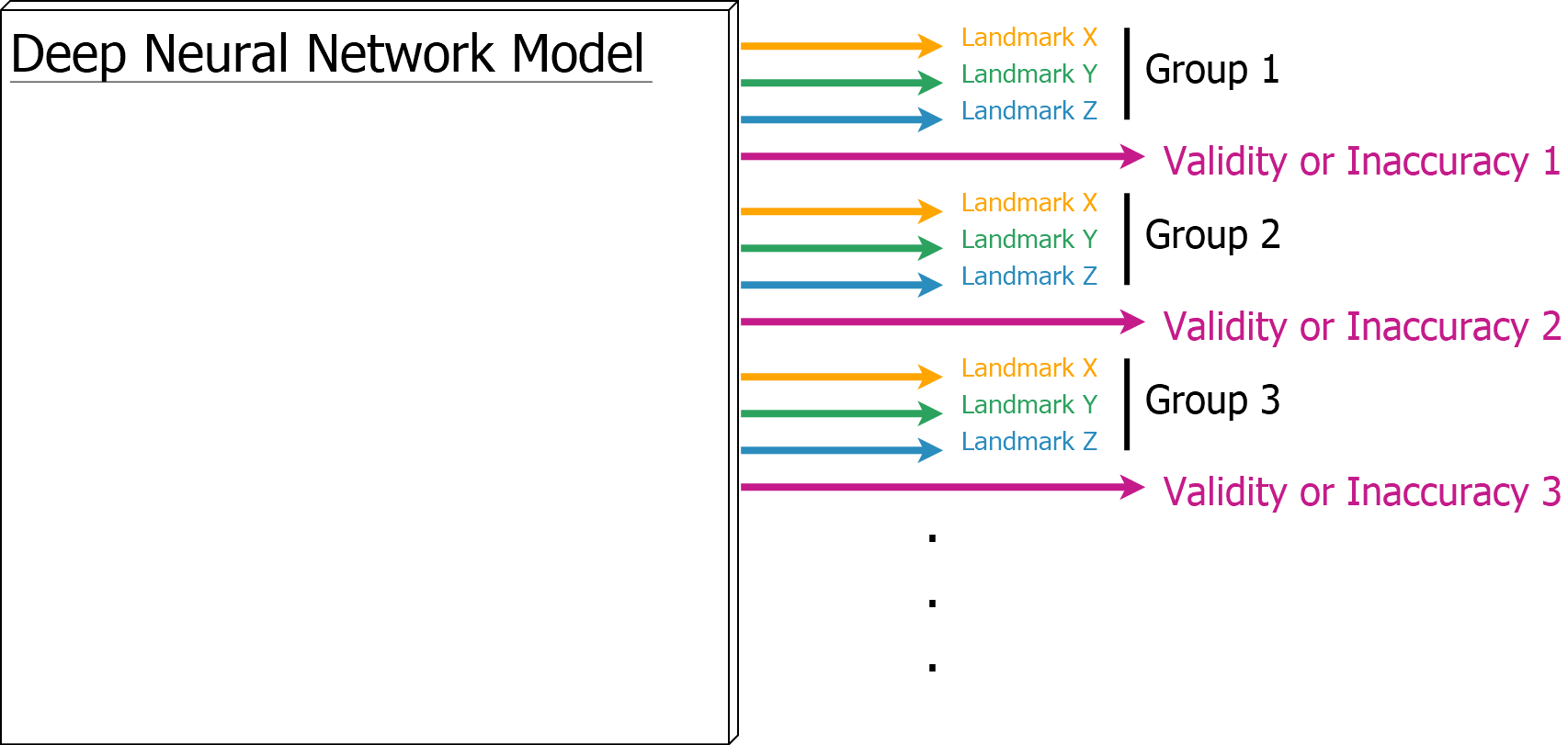}
		\caption{Integration of the inaccuracy into deep neural network architectures. Each landmark and its diemensons forms one group and gets one additional neuron as output which estimates the inaccuracy.}
		\label{fig:dnnstructure}
	\end{figure}
	
	The proposed approach is an extension of the validation loss from \cite{fuhl2020learning}. In the original formulation (Equation~\ref{eq:Original}) the authors proposed to add an additional output neuron per landmark. This additional neuron learns to approximate the error of the estimated landmark it belongs too. Therefore, each landmark forms a group, as can be seen in Figure~\ref{fig:dnnstructure}. For each group, so each landmark, an additional neuron is added and drawn as a pink arrow in Figure~\ref{fig:dnnstructure}. Based on Equation~\ref{eq:Original} this neuron estimates the sum of absolute errors. The disadvantage of this kind of formulation is that the inaccuracy is not normalized based on the area of the object nor the amount of landmarks. This means that the inaccuracy signal produces a higher gradient in contrast to the landmark signals itself, since it is numerically larger. Therefore, the neural network puts more effort into learning a correct validation or inaccuracy signal compared to the correct estimation of the landmarks. This disadvantage lead to our idea of adding a normalization to Equation~\ref{eq:Original} which can be seen in Equation~\ref{eq:AbsInaccuracy}.
	
	\begin{equation}
		Loss_{Inaccuracy}=((\frac{\sum_{i=0}^{GroupSize} |GT_{i}-ES_{i}|}{\frac{AREA}{min(AREA)}*GroupSize}) - ES_{Inaccuracy})^2
		\label{eq:AbsInaccuracy}
	\end{equation}
	
	$GroupSize$ in Equation~\ref{eq:AbsInaccuracy} refers to the amount of dimensions for one landmark, as can be seen in Figure~\ref{fig:dnnstructure}. $GT_{i}$ is the ground truth for a landmark at the ith position in the group, and $ES_{i}$ is the estimated landmark position of the ith neuron in the group. $\frac{AREA}{min(AREA)}$ is the normalization factor for the estimated complete shape of all landmarks. $AREA$ is the amount of pixels of the shape at hand and $min(AREA)$ is the amount of pixels of the smalles shape in the image at hand. The idea behind this normalization part is that larger shapes do not receive higher gradients compared to smaller shapes. This is for example the case in Pistol~\cite{fuhl2022pistol} where the Iris and the eyelid estimates produces much higher validation or inaccuracy errors compared to the pupil estimates. We use the pixelwise relation since we do not know the shape in advance, and we want that our approach applies to as many shapes as possible without modification. Additionally, to the normalization factor of the shape size ($\frac{AREA}{min(AREA)}$) we use the group size ($GroupSize$) in the denominator. This factor normalizes the gradient to be numerically fair in the group for each landmark.
	
	\begin{equation}
		Loss_{Inaccuracy}=((\frac{\sqrt{\sum_{i=0}^{GroupSize} (GT_{i}-ES_{i})^{2}}}{\frac{AREA}{min(AREA)}}) - ES_{Inaccuracy})^2
		\label{eq:EuclidInaccuracy}
	\end{equation}
	
	As an alternative to the absolute error formulation as shown in Equation~\ref{eq:AbsInaccuracy} we also formulated the validation or inaccuracy loss as Euclidean error (Equation~\ref{eq:EuclidInaccuracy}). Here we did not add the group size to the denominator since the Euclidean norm already normalizes the numerical output. All variables have the same meaning as described for Equation~\ref{eq:AbsInaccuracy}, the only difference is the Euclidean norm.
	
	\begin{equation}
		Gradient = \left\{
		\begin{array}{ll}
			Gradient_{Inaccuracy}*LearningRate & |Gradient_{Inaccuracy}| \ge Margin \\
			0 & \, \textrm{else} \\
		\end{array}
		\right. 
		\label{eq:Margin}
	\end{equation}
	
	Equation~\ref{eq:Margin} describes our gradient masking to improve the accuracy of the trained neural networks. The idea behind the gradient masking is that small gradients come from either numerical instabilities of the floating point representation or from inaccurately annotated landmarks. Therefore, we want to suppress them during the training phase, and this is done by a threshold of the gradient called $Margin$ in Equation~\ref{eq:Margin}. We compare this $Margin$ with the absolute value of the gradient and if it is in this $Margin$, it is set to zero, otherwise we use the gradient for learning.
	
	\section{Evaluation}
	In this section we describe the used dataset, the training parameters and report our results. The first part will be the training parameters and the dataset description. Afterward, we perform an initial parameter evaluation to select an optimal margin for Equation~\ref{eq:Margin} based on the training data only. The results for the margin evaluation can be seen in Table~\ref{tbl:margineval}. In the following, we compare our approach to the original formulation of the loss function.
	
	\subsection{Dataset}
	\label{ds:split}
	We used the real world data from the TEyEDS dataset~\cite{fuhl2021teyed}. Since the entire dataset is too huge for our hardware capabilities, we selected a subset of 120 Dikablis videos for training, 30 Dikablis Videos for validation and additionally 50 videos for testing. For each video we extracted every 10th frame since many consecutive images are very similar in eye tracking recordings due to the high sampling rate and the immovable eye during fixations. The Dikablis videos from the TEyEDS dataset~\cite{fuhl2021teyed} are real world driving videos from a study conducted in Tübingen. The data contains a lot of reflections on glasses, make up as well as eye tracker shifts and bad illuminated eye images. This makes it especially challenging and increase the value of our results since the recorded images are from real world settings without any limitation or restrictions to the participants. 
	
	\subsection{Training parameters}
	For training of all models we used a batch size of 10, initial learning rate of $10^{-4}$ and the Adam optimizer~\cite{kingma2014adam}. We set the first momentum to 0.9, the second momentum to 0.999 and the weight decay to $5*10^{-4}$. As data augmentation we used adding reflections, adding noise, eye image shifts, and deformations of the image itself as well as for the annotations. We trained each model for 50 epochs and reduced the learning rate from $10^{-4}$ to $10^{-5}$. With the learning rate of $10^{-5}$ we continued the training for additionally 50 epochs and used the best model based on the validation set results. The margin parameter from Equation~\ref{eq:Margin} was set to 0.005 after the evaluation on the training set shown in Table~\ref{tbl:margineval}.

	\subsection{Results}
	
	\begin{table*}
		\centering
		\caption{ Results for different margins and different models averaged over 5 runs. MIoU is the mean intersection over union and MED is the mean euclidean distance of the landmarks. For the evaluation we used 50 videos from TEyEDS~\cite{fuhl2021teyed} for training, 20 videos for validation and 50 videos for testing. \textbf{This split was done on the training data only as described in Section~\ref{ds:split}.}}
		\label{tbl:margineval}
		\begin{tabular}{lccccccc}
			&  & \multicolumn{2}{c}{Pupil} & \multicolumn{2}{c}{Iris}& \multicolumn{2}{c}{Eyelid}\\ 
			Model & Margin & MIoU & MED & MIoU & MED & MIoU & MED \\ \hline
			\multirow{6}{*}{ResNeXt-18~\cite{xie2017aggregated} + EQ~\ref{eq:AbsInaccuracy}} & 0.001 & 0.71 & 1.45 & 0.70 & 1.68 & 0.70 & 2.03 \\ 
			& 0.005 & 0.73 & 1.34 & 0.72 & 1.59 & 0.72 & 1.95 \\
			& 0.01 & 0.72 & 1.40 & 0.71 & 1.62 & 0.70 & 2.06 \\ 
			& 0.02 & 0.69 & 1.54 & 0.68 & 1.74 & 0.66 & 2.32 \\ 
			& 0.04 & 0.61 & 1.68 & 0.61 & 1.89 & 0.64 & 2.44 \\ 
			& 0.05 & 0.59 & 1.72 & 0.56 & 1.97 & 0.61 & 2.61 \\ \hline		
			\multirow{6}{*}{ResNeXt-18~\cite{xie2017aggregated} + EQ~\ref{eq:EuclidInaccuracy}} & 0.001 & 0.72 & 1.34 & 0.71 & 1.61 & 0.71 & 2.01 \\ 
			& 0.005 & 0.73 & 1.29 & 0.73 & 1.56 & 0.72 & 1.93 \\
			& 0.01 & 0.71 & 1.48 & 0.70 & 1.69 & 0.69 & 2.13\\ 
			& 0.02 & 0.69 & 1.55 & 0.67 & 1.77 & 0.66 & 2.31 \\ 
			& 0.04 & 0.62 & 1.61 & 0.60 & 1.88 & 0.63 & 2.46 \\ 
			& 0.05 & 0.59 & 1.69 & 0.55 & 2.08 & 0.61 & 2.58 \\ \hline
		\end{tabular}
	\end{table*}
	
	For the selection of the optimal margin parameter from Equation~\ref{eq:Margin} we used the training set and split it into 50 videos to train, 20 videos for validation and 50 videos for testing. We evaluated different margins, as can be seen in Table~\ref{tbl:margineval} with the ResNeXt model and both equations for the inaccuracy estimation. The best margin for both equations was 0.005 which we used for all the following evaluations. As can be seen in Table~\ref{tbl:margineval}, the margin parameter is more or less stable in the area around 0.001 to 0.01, but afterward it has a huge impact on the model. This is the case since important gradients are blocked and hinder the network from learning by jumping from no gradient to a high gradient if the error increases. Since the gradient is dependent on the numerical stability of floating point values and the accuracy of the dataset annotations, this margin has to be reevaluated if less accurately annotated datasets are used. Overall it has to be said, that we did not perform a very fine-grained selection of the margin parameter and based on the values between 0.001 and 0.01, there are possibly other values which would improve the results of the model further. We just want to show the impact of the parameter on the model, especially if it is set too high. In addition, we wanted to perform a coarse evaluation in the useful range (Below 0.01) to justify our selection of the margin parameter.
	
	\begin{table*}
		\centering
		\caption{ Results for different models on the test set averaged over 5 runs. Here we evaluate the impact of the new normalization together with the margin on the accuracy of the extracted shape. As metric we used MIoU which is the mean intersection over union. In addition, we used the MED metric which is the mean euclidean distance of the landmarks. As state-of-the-art we used the original approach from \cite{fuhl2020learning}.}
		\label{tbl:accuracyeval}
		\begin{tabular}{llcccccc}
			& &  \multicolumn{2}{c}{Pupil} & \multicolumn{2}{c}{Iris}& \multicolumn{2}{c}{Eyelid}\\ 
			& Model &  MIoU & MED & MIoU & MED & MIoU & MED \\ \hline
			\multirow{6}{*}{\rotatebox[origin=c]{90}{State-of-the-art}} & ResNeXt-18~\cite{xie2017aggregated} + \cite{fuhl2020learning} & 0.74 & 1.27 & 0.72 & 1.60 & 0.71 & 2.01 \\ 
			& ResNeXt-32~\cite{xie2017aggregated} + \cite{fuhl2020learning} & 0.74 & 1.25 & 0.73 & 1.55 & 0.71 & 1.98 \\ 
			& ResNeXt-50~\cite{xie2017aggregated} + \cite{fuhl2020learning} & 0.74 & 1.24 & 0.73 & 1.54 & 0.71 & 1.97 \\ 
			& InceptionV4~\cite{szegedy2017inception} + \cite{fuhl2020learning} & 0.74 & 1.21 & 0.73 & 1.51 & 0.72 & 1.94 \\ 
			& EfficentNet~\cite{tan2019efficientnet} + \cite{fuhl2020learning} & 0.74 & 1.25 & 0.73 & 1.56 & 0.71 & 1.97\\ 
			& PyramidNet~\cite{han2017deep} + \cite{fuhl2020learning} & 0.74 & 1.23 & 0.73 & 1.52 & 0.71 & 1.95 \\ \hline
			\multirow{6}{*}{\rotatebox[origin=c]{90}{Proposed}} & ResNeXt-18~\cite{xie2017aggregated} + EQ~\ref{eq:AbsInaccuracy} & 0.76 & 1.14 & 0.73 & 1.57 & 0.73 & 1.88 \\ 
			& ResNeXt-32~\cite{xie2017aggregated} + EQ~\ref{eq:AbsInaccuracy} & 0.77 & 1.08 & 0.74 & 1.53 & 0.73 & 1.84 \\ 
			& ResNeXt-50~\cite{xie2017aggregated} + EQ~\ref{eq:AbsInaccuracy} & 0.77 & 1.06 & 0.75 & 1.48 & 0.73 & 1.81 \\ 
			& InceptionV4~\cite{szegedy2017inception} + EQ~\ref{eq:AbsInaccuracy} & 0.77 & 1.03 & 0.75 & 1.42 & 0.74 & 1.62 \\ 
			& EfficentNet~\cite{tan2019efficientnet} + EQ~\ref{eq:AbsInaccuracy} & 0.77 & 1.07 & 0.75 & 1.50 & 0.73 & 1.83\\ 
			& PyramidNet~\cite{han2017deep} + EQ~\ref{eq:AbsInaccuracy} & 0.77 & 1.05 & 0.75 & 1.44 & 0.73 & 1.80\\ \hline		
			\multirow{6}{*}{\rotatebox[origin=c]{90}{Proposed}} & ResNeXt-18~\cite{xie2017aggregated} + EQ~\ref{eq:EuclidInaccuracy} & 0.77 & 1.10 & 0.74 & 1.53 & 0.73 & 1.89 \\ 
			& ResNeXt-32~\cite{xie2017aggregated} + EQ~\ref{eq:EuclidInaccuracy} & 0.77 & 1.09 & 0.74 & 1.52 & 0.73 & 1.87 \\ 
			& ResNeXt-50~\cite{xie2017aggregated} + EQ~\ref{eq:EuclidInaccuracy} & 0.78 & 1.01 & 0.75 & 1.46 & 0.74 & 1.70 \\ 
			& InceptionV4~\cite{szegedy2017inception} + EQ~\ref{eq:EuclidInaccuracy} & 0.78 & 0.96 & 0.75 & 1.38 & 0.75 & 1.56 \\ 
			& EfficentNet~\cite{tan2019efficientnet} + EQ~\ref{eq:EuclidInaccuracy} & 0.78 & 1.03 & 0.75 & 1.48 & 0.74 & 1.72 \\ 
			& PyramidNet~\cite{han2017deep} + EQ~\ref{eq:EuclidInaccuracy} & 0.78 & 1.00 & 0.75 & 1.45 & 0.74 & 1.69 \\ \hline
		\end{tabular}
	\end{table*}
	
	Table~\ref{tbl:accuracyeval} shows the impact to the accuracy compared to the original approach. The first section in Table~\ref{tbl:accuracyeval} are the results from the original approach. As can be seen for the different ResNeXt models, the accuracy only improves slightly for larger models. Compared to the normalized version together with the margin, the ResNeXt models have a higher increase in performance with an increase of the model size. This is the case for Equation~\ref{eq:AbsInaccuracy} and Equation~\ref{eq:EuclidInaccuracy}. In addition, the new formulation of the loss function and the margin increase the performance of all models significantly. This is the case for all models and all tasks (Pupil, Iris, and Eyelid landmark estimation). If we compare the models for Equation~\ref{eq:AbsInaccuracy} and Equation~\ref{eq:EuclidInaccuracy} we see that Equation~\ref{eq:EuclidInaccuracy} slightly outperforms Equation~\ref{eq:AbsInaccuracy}. This is possibly due to the fact that the landmarks are from a Euclidean space, where the Euclidean norm applies better compared to the averaged absolute norm. While it is here the case that Equation~\ref{eq:EuclidInaccuracy} is superior, this could change based on other datasets or landmarks from non-Euclidean geometries. 
	
	\begin{table*}
		\centering
		\caption{ This evaluation shows the evaluation for the exclusion of inaccurate landmarks which is also mentioned as correction in this paper. As state-of-the-art we used the original approach from \cite{fuhl2020learning}. \textbf{The selected exclusion threashold was computed on the validation set and the optimal threashold based on the MIoU and MED score was selected for each model seperately.} MIoU is the mean intersection over union and MED is the mean euclidean distance of the landmarks. The evaluation dataset is a subset selected from TEyEDS~\cite{fuhl2021teyed} from the Dikablis eye tracker videos. Each model was trained and evaluated five times and we report the average performance on the test set here.}
		\label{tbl:correction}
		\begin{tabular}{llcccccc}
			& &  \multicolumn{2}{c}{Pupil} & \multicolumn{2}{c}{Iris}& \multicolumn{2}{c}{Eyelid}\\ 
			& Model &  MIoU & MED & MIoU & MED & MIoU & MED \\ \hline
			\multirow{6}{*}{\rotatebox[origin=c]{90}{State-of-the-art}} & ResNeXt-18~\cite{xie2017aggregated} + \cite{fuhl2020learning} & 0.84 & 0.83 & 0.79 & 1.43 & 0.73 & 1.83 \\ 
			& ResNeXt-32~\cite{xie2017aggregated} + \cite{fuhl2020learning} & 0.84 & 0.81 & 0.80 & 1.34 & 0.74 & 1.72 \\ 
			& ResNeXt-50~\cite{xie2017aggregated} + \cite{fuhl2020learning} & 0.85 & 0.79 & 0.80 & 1.31 & 0.75 & 1.60 \\ 
			& InceptionV4~\cite{szegedy2017inception} + \cite{fuhl2020learning} & 0.85 & 0.76 & 0.80 & 1.29 & 0.75 & 1.59 \\ 
			& EfficentNet~\cite{tan2019efficientnet} + \cite{fuhl2020learning} & 0.85 & 0.79 & 0.80 & 1.33 & 0.75 & 1.61\\ 
			& PyramidNet~\cite{han2017deep} + \cite{fuhl2020learning} & 0.85 & 0.81 & 0.80 & 1.32 & 0.75 & 1.61 \\ \hline		
			\multirow{6}{*}{\rotatebox[origin=c]{90}{Proposed}} & ResNeXt-18~\cite{xie2017aggregated} + EQ~\ref{eq:AbsInaccuracy} & 0.90 & 0.64 & 0.87 & 1.04 & 0.84 & 1.38 \\ 
			& ResNeXt-32~\cite{xie2017aggregated} + EQ~\ref{eq:AbsInaccuracy} & 0.91 & 0.55 & 0.88 & 0.97 & 0.84 & 1.21 \\ 
			& ResNeXt-50~\cite{xie2017aggregated} + EQ~\ref{eq:AbsInaccuracy} & 0.92 & 0.53 & 0.88 & 0.93 & 0.84 & 1.20 \\ 
			& InceptionV4~\cite{szegedy2017inception} + EQ~\ref{eq:AbsInaccuracy} & 0.92 & 0.52 & 0.88 & 0.91 & 0.84 & 1.20 \\ 
			& EfficentNet~\cite{tan2019efficientnet} + EQ~\ref{eq:AbsInaccuracy} & 0.92 & 0.56 & 0.88 & 0.95 & 0.84 & 1.24\\ 
			& PyramidNet~\cite{han2017deep} + EQ~\ref{eq:AbsInaccuracy} & 0.92 & 0.54 & 0.88 & 0.94 & 0.84 & 1.23\\ \hline		
			\multirow{6}{*}{\rotatebox[origin=c]{90}{Proposed}} & ResNeXt-18~\cite{xie2017aggregated} + EQ~\ref{eq:EuclidInaccuracy} & 0.91 & 0.57 & 0.88 & 1.02 & 0.84 & 1.36 \\ 
			& ResNeXt-32~\cite{xie2017aggregated} + EQ~\ref{eq:EuclidInaccuracy} & 0.91 & 0.55 & 0.88 & 0.92 & 0.84 & 1.22 \\ 
			& ResNeXt-50~\cite{xie2017aggregated} + EQ~\ref{eq:EuclidInaccuracy} & 0.92 & 0.51 & 0.89 & 0.84 & 0.85 & 1.13 \\ 
			& InceptionV4~\cite{szegedy2017inception} + EQ~\ref{eq:EuclidInaccuracy} & 0.92 & 0.51 & 0.89 & 0.82 & 0.85 & 1.11 \\ 
			& EfficentNet~\cite{tan2019efficientnet} + EQ~\ref{eq:EuclidInaccuracy} & 0.92 & 0.54 & 0.89 & 0.88 & 0.85 & 1.16 \\ 
			& PyramidNet~\cite{han2017deep} + EQ~\ref{eq:EuclidInaccuracy} & 0.92 & 0.53 & 0.89 & 0.86 & 0.85 & 1.15 \\ \hline
		\end{tabular}
	\end{table*}
	
	Table~\ref{tbl:correction} shows the results of all models if we use the inaccuracy signal to perform an outlier selection of the landmarks for shape estimation. If we compare Table~\ref{tbl:accuracyeval} with Table~\ref{tbl:correction} it is clear to see that all model receive an improved performance. This means all inaccuracy signals are meaningful. If we take a look onto the increase in performance between Table~\ref{tbl:accuracyeval} and Table~\ref{tbl:correction}, the original loss formulation has an increase of 0.10 for the pupil MIoU, 0.07 for the iris MIoU, and 0.03 for the eyelid MIoU. The models based on Equation~\ref{eq:AbsInaccuracy} increase their performance by 0.15, 0.13, and 0.11 of the MIoU for pupil, iris, and eyelid in the respective order. While this increase outperforms the original formulation, it is also obvious that larger areas like the iris and the eyelid also gain a large boost. This is mainly due to the normalization of the area of the entire shape ($\frac{AREA}{min(AREA)}$ in Equation~\ref{eq:AbsInaccuracy} and Equation~\ref{eq:EuclidInaccuracy}) since this normalization factor reduces the higher error of larger shapes to suppress the errors from smaller shapes. Therefore, the inaccuracy estimation is more similar and allows a more stable outlier selection. For Equation~\ref{eq:EuclidInaccuracy} the improvements between Table~\ref{tbl:accuracyeval} and Table~\ref{tbl:correction} are 0.14, 0.14, and 0.1 for the MIoU score on the pupil, iris, and eyelid respectively. Those improvements are similar compared to Equation~\ref{eq:AbsInaccuracy} which confirms our hypothesis that the normalization by the area relation is the main contributor here. Overall, the models trained with the loss function from Equation~\ref{eq:EuclidInaccuracy} outperform the models from the original formulation as well as the models from Equation~\ref{eq:AbsInaccuracy} which is possibly due to the Euclidean space the landmarks are from.
	
	\section{Limitations}
	We tried to include many different deep learning architecutres into our evaluation, but we did not include vison transformers~\cite{dosovitskiy2020image}. The first reason why we did not include vision transformers is that the training and batch size requirements are larger compared to conventional convolutional architectures. Therefore, all of our evaluations would have required much more energy as well as much more trining time. In addition, we would need to use possibly more than the entire  TEyeDS~\cite{fuhl2021teyed} dataset to bring the transformers to a level where they really beat convolutional approaches. The second reason why we did not use transformers is simple and are our hardware limitations. We have access to exactly four 3090 GPUs from NVIDIA which would only have enough memory if we use them together. For the concolutional nets in contrast we could train each model seperately on one GPU.\\
	Another limitation of our evaluation is that we only used eye images for training. We did not show that our approach works for facial images or any other landmark based application. We expect that it works for any kind of landmark based approach, but as mentioned before, we did not explicitly show that in our evaluation. We did not include such evaluations due to the page limit and our specific application case of eye tracking feature extraction.
	
	\section{Conclusion and Outlook}
	In this paper, we proposed an extension to the validation loss formulation from \cite{fuhl2020learning}. We showed the impact of the margin parameter on the model performance in Table~\ref{tbl:margineval} and evaluated the impact of our extension to the model performance in Table~\ref{tbl:accuracyeval}. The extension outperformed the original formulation of the validation loss function for all models and all tasks, and shows a significant improvement. In Table~\ref{tbl:correction} we evaluated the usefulness of the validation or inaccuracy signal for outlier detection. Here our extension has the larges impact between the different shapes which the models have to estimate based on the landmarks (Pupil, iris, and eyelid). This can be seen by comparing the performance improvement between Table~\ref{tbl:correction} and Table~\ref{tbl:accuracyeval} for the original and the extended loss formulation. In addition, the overall model performance improves significantly together with the outlier detection for both proposed loss formulations (Equation~\ref{eq:AbsInaccuracy} and Equation~\ref{eq:EuclidInaccuracy}).
	
	Further research should investigate the impact of both equations (\ref{eq:AbsInaccuracy} and \ref{eq:EuclidInaccuracy}) for other datasets and especially for non-Euclidean geometries. This could lead to a more general formulation of the loss function which works well in different geometries. Since the main impact for the outlier detection results (Table~\ref{tbl:correction}) came from the normalization based on the shape, which is estimated based on the landmarks, further research should investigate if there are more suitable and general formulations for this kind of normalization. 
	
	\section{Potentially harmful impacts and future societal risks}
	While the proposed approach is mainly directed into the research area of eye tracking, it could be trained for anything like enemy soldier detection and segmentation, monitoring of civilians, quality control in the process of weapon manufacturing, and possibly many more. Therefore, it is possible that people use such findings in application areas that are criminal or at least ethically difficult to comprehend. We as researchers do not want our knowledge to be used in such areas, but we cannot prevent it from happening.

	\section*{Acknowledgement}
	Funded by the Deutsche Forschungsgemeinschaft (DFG, German Research Foundation) – 508330921
	
	\bibliographystyle{plain}
	\bibliography{template}

\end{document}